\documentclass[10pt,twocolumn,letterpaper]{article}

\usepackage[pagenumbers]{cvpr}  

\definecolor{cvprblue}{rgb}{0.21,0.49,0.74}
\usepackage[pagebackref,breaklinks,colorlinks,allcolors=cvprblue]{hyperref}
\usepackage{booktabs}
\usepackage{multirow}
\usepackage{colortbl}      
\definecolor{afssrow}{gray}{0.92}  
\usepackage{arydshln}
\usepackage{tabularx}
\usepackage{ragged2e}
\def\paperID{41317} 
\def\confName{CVPR}
\def\confYear{2026}

\title{Does YOLO Really Need to See Every Training Image in Every Epoch?}

\author{
	Xingxing Xie$^{1}$
	\and 
	Jiahua Dong$^{1}$
	\and
	Junwei Han$^{1,2}$
	\and
	Gong Cheng$^{1*}$
	\and 
	$^{1}$ School of Automation, Northwestern Polytechnical University \\
	$^{2}$ School of Artificial Intelligence, Chongqing University of Posts and Telecommunications \\
    {\tt\small \{xiexing,gcheng,jhan\}@nwpu.edu.cn} \hspace{5mm} {\tt\small jiahuadong@mail.nwpu.edu.cn}
}

\begin{document}
	\twocolumn[{%
		\renewcommand\twocolumn[1][]{#1}%
		\maketitle
		\begin{center}
			\centering
			\includegraphics[width=1\textwidth]{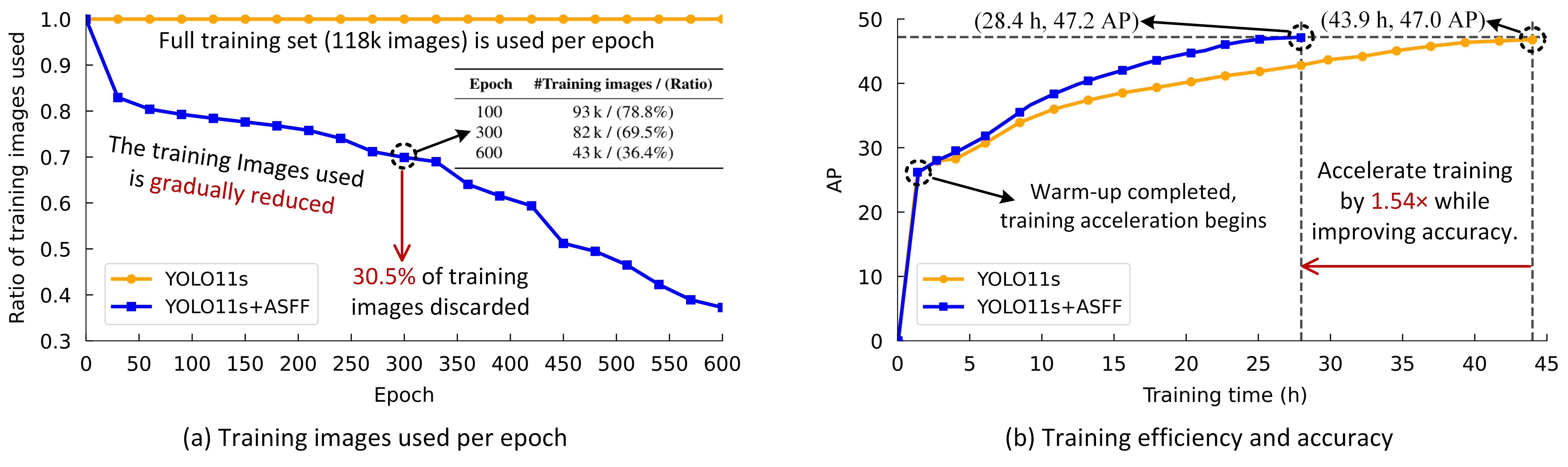}
			\vspace*{-6mm}
			\captionof{figure}{Comparison of YOLO11s with and without AFSS in terms of the training images used and the corresponding training efficiency and accuracy on MS COCO 2017. (a) Training data used per epoch: AFSS adaptively selects the images used for training, progressively reducing the number of images utilized over time, whereas the vanilla YOLO11s employs the full training set in every epoch; (b) Training efficiency and accuracy: AFSS accelerates YOLO11s training by 1.54 $\times$ while improving detection accuracy.}
			\label{fig1}
			\vspace{0.5\baselineskip}
		\end{center}%
	}]

\maketitle
\renewcommand{\thefootnote}{\fnsymbol{footnote}}
\footnotetext{*Gong Cheng is corresponding author.}

\begin{abstract}
YOLO detectors are known for their fast inference speed, yet training them remains unexpectedly time-consuming due to their exhaustive pipeline that processes every training image in every epoch, even when many images have already been sufficiently learned. This stands in clear contrast to the efficiency suggested by the ``You Only Look Once'' philosophy.
This naturally raises an important question: \textit{Does YOLO really need to see every training image in every epoch?} To explore this, we propose an Anti-Forgetting Sampling Strategy (AFSS) that dynamically determines which images should be used and which can be skipped during each epoch, allowing the detector to learn more effectively and efficiently.
Specifically, AFSS measures the learning sufficiency of each training image as the minimum of its detection recall and precision, and dynamically categorizes training images into easy, medium, or hard levels accordingly. Easy training images are sparsely resampled during training in a continuous review manner, with priority given to those that have not been used for a long time to reduce redundancy and prevent forgetting. Moderate training images are partially selected, prioritizing recently unused ones and randomly choosing the rest from unselected images to ensure coverage and prevent forgetting. Hard training images are fully sampled in every epoch to ensure sufficient learning. The learning sufficiency of each training image is periodically updated, enabling detectors to adaptively shift its focus toward the informative training images over time while progressively discarding redundant ones.
On widely used natural image detection benchmarks (MS COCO 2017 and PASCAL VOC 2007) and remote sensing detection datasets (DOTA-v1.0 and DIOR-R), AFSS achieves more than $1.43\times$ training speedup for YOLO-series detectors while also improving accuracy.

\end{abstract}    
\section{Introduction}
\label{sec:intro}

Object detection is one of the cornerstone tasks in computer vision and serves as a fundamental capability that enables intelligent perception systems to understand and interpret the real world~\cite{zou2023object,liu2020deep,ren2016faster,lin2017focal,carion2020end,madet}. It underpins a wide range of applications such as autonomous driving~\cite{almalioglu2022deep,hu2023planning}, robotics~\cite{liu2022behavior,ding2023learning}, surveillance~\cite{otgonbold2024simple,fu2019foreground}, and medical image analysis~\cite{rui2025multi,jang2022m3t}. Rapid progress in deep learning has driven substantial advancements in object detection, improving both speed and accuracy. Among existing detection approaches~\cite{liu2016ssd,DFNet,law2018cornernet,carion2020end,ORCNN}, YOLO (You Only Look Once) family~\cite{redmon2016you,redmon2017yolo9000,redmon2018yolov3,wang2023yolov7,wang2023gold,wang2024yolov10,tian2025yolov12} has emerged as one of the most influential paradigms, primarily owing to its exceptional inference efficiency. Since the release of YOLOv1~\cite{redmon2016you}, the series has undergone continuous architectural and algorithmic evolution, transitioning from a unified regression framework to modern designs that incorporate multi-scale feature fusion~\cite{liu2018path,lin2017feature}, decoupled detection heads~\cite{ge2021yolox,feng2021tood}, dynamic label assignment~\cite{ge2021ota,zhang2020bridging}, and lightweight backbones~\cite{he2016deep,ding2021repvgg}. In particular, recent versions such as YOLOv8 through YOLO12 have further advanced the frontier of real-time detection, achieving unprecedented inference speed and higher accuracy. As a result, YOLO has become the de facto standard for real-time detection systems across both research and industry~\cite{wang2024yolov10,tian2025yolov12,sapkota2025yolo26}.

However, this seemingly perfect efficiency conceals an intriguing paradox: YOLO demonstrates outstanding inference capability, yet its training remains surprisingly time-consuming. For example, on the MS COCO 2017 dataset~\cite{lin2014microsoft}, lightweight YOLO11s achieves an inference speed of 200 FPS, with training taking 43.9 hours on a configuration with two RTX 4090 GPUs. In contrast, under the same conditions, Faster R-CNN with ResNet50 requires only 6.5 hours to train, which is 6.9 times faster than YOLO11s. This striking asymmetry between inference and training reveals a fundamental inconsistency: a model built on the philosophy of ``You Only Look Once'' during inference still requires prolonged training. Such a gap suggests that structural efficiency does not necessarily translate into efficient learning.

To better understand this inefficiency, we revisit the training behavior of the YOLO family from the perspective of training image utilization. As illustrated in Fig.~\ref{fig1}(a), YOLO adopts a full-coverage training paradigm in which the entire training set is traversed in every epoch, and each image participates in both forward and backward passes. In other words, every training image is processed hundreds of times throughout the training procedure. Although this exhaustive scheme ensures complete data exposure and stable early optimization~\cite{he2019rethinking}, it lacks adaptability to the evolving learning dynamics of the model. Once the model has sufficiently learned from certain training images, continuing to process them at the same frequency yields diminishing returns while consuming substantial training time, as shown in Fig.~\ref{fig1}(b). Over hundreds of epochs, such redundant computation accumulates over time, turning the training process into a prolonged and mechanically repetitive routine that prioritizes completeness over necessity. More fundamentally, this paradigm implicitly assumes that all training images contribute equally throughout training~\cite{wang2024yolov10,tian2025yolov12}, an assumption at odds with representation learning, since the informativeness of each training image naturally changes as the model evolves~\cite{kleinman2021usable,katharopoulos2018not}. A truly efficient learning system should therefore adapt its data usage over time, allocating computation to the most informative training images while preserving previously acquired knowledge. This observation leads to a central question underlying YOLO's training inefficiency: Does YOLO really need to see every training image in every epoch? If not, can a detector learn more efficiently by dynamically selecting which training images to revisit and which to skip, without forgetting what has already been learned?

Motivated by the observation that YOLO detectors repeatedly process well-learned training images, we propose an Anti-Forgetting Sampling Strategy (AFSS) to make training adaptive rather than uniform. AFSS adjusts the participation frequency of each training image based on its learning sufficiency while preventing the forgetting of acquired knowledge. Learning sufficiency measures how well each training image has been learned and is estimated from the precision-recall statistics of the previous epoch. Based on this measure, training images are categorized as easy, moderate, or hard, with each group following a tailored sampling policy. Easy training images are sampled with very low frequency to reduce redundant computation, while those that have not been visited for an extended period are periodically resampled to preserve their learned representations. Moderate training images are selected at a fixed proportion, and recently unused training images are forcibly resampled to maintain coverage and optimization stability. Hard training images are included in every epoch to ensure continuous learning of challenging cases. As the model evolves, AFSS continuously updates the learning sufficiency score of each training image, enabling the data distribution presented to the model to co-evolve with its internal representation.

On large-scale and widely used natural image object detection benchmarks such as MS~COCO~2017~\cite{lin2014microsoft} and PASCAL~VOC~2007~\cite{everingham2015pascal}, AFSS accelerates the training of YOLO-based detectors (YOLOv8, YOLOv10, YOLO11, and YOLO12) by more than 1.43$\times$ without sacrificing accuracy. For oriented object detection in remote sensing images, we evaluate AFSS on the DOTA-v1.0~\cite{xia2018dota} and DIOR-R~\cite{cheng2022anchor} datasets using YOLOv8-OBB and YOLO11-OBB, which are widely adopted and representative models. AFSS accelerates their training by over 1.63$\times$ while maintaining accuracy. These results demonstrate that efficient learning does not require exhaustive repetition. By teaching YOLO \emph{what} to see and \emph{when} to see it, AFSS effectively bridges the gap between fast perception and efficient learning.

\section{Related Work}
Improving training efficiency of object detectors without altering their architecture has emerged as an increasingly important yet challenging direction~\cite{bengio2009curriculum,kumar2010self,paul2021deep,lei2023comprehensive}. Existing studies can be roughly grouped into three lines of research: curriculum and self-paced learning, dataset pruning and dataset distillation.

\textbf{Curriculum and self-paced learning}~\cite{bengio2009curriculum,kumar2010self} organize data according to sample difficulty, starting with easy image samples and gradually introducing harder ones. Some works~\cite{sangineto2018self, soviany2021curriculum} integrate these ideas into object detection by iteratively estimating the reliability of training images and gradually expanding the selected samples from easy to hard.
This strategy stabilizes early optimization but prevents difficult images from participating throughout training. As a result, models learn insufficiently from complex cases, limiting robustness and final accuracy. Such methods optimize the order of learning but fail to ensure consistent exposure to informative images.

\textbf{Dataset pruning}~\cite{paul2021deep,yang2023dataset,toneva2018an} accelerates training by removing redundant or low-value images. Representative work such as Deep Learning on a Data Diet~\cite{paul2021deep} selects important images using early loss or gradient statistics and discards the rest. While pruning reduces computational cost, it inevitably introduces information forgetting and bias. Once removed, image samples can no longer be revisited, causing the model to forget certain patterns and overfit to the retained subset. This static and irreversible process lacks adaptability and is particularly problematic for long-epoch detectors such as YOLO.

\textbf{Dataset distillation}~\cite{lei2023comprehensive,qi2024fetch} further reduces training cost by synthesizing a small set of artificial images that approximate the behavior of the original dataset. Recent work such as Fetch and Forge~\cite{qi2024fetch} extends this idea to detection, creating representative synthetic images to replace real data. However, this paradigm requires additional data synthesis through complex bi-level optimization and generative modeling. The produced samples often fail to capture real-world noise, occlusion, and context, leading to reduced accuracy and limited scalability.

Unlike curriculum and self-paced learning, which gradually introduce hard images and fail to maintain their continuous participation, our Anti-Forgetting Sampling Strategy (AFSS) keeps hard images engaged throughout entire training process to ensure sufficient learning.
In contrast to dataset pruning, which discards images deemed less valuable and is prone to forgetting and bias, AFSS integrates an anti-forgetting mechanism that periodically revisits well-learned easy images and maintains short-term coverage of moderate ones to preserve stability and prevent knowledge degradation.
Different from dataset distillation, which relies on synthetic data and struggles to maintain image diversity, AFSS achieves full coverage of hard images, periodic revisiting of easy ones, and short-term coverage of moderate ones, leading to a more diverse and balanced training process. A comparison with these methods is listed in Table~\ref{tab3}.


\section{Method}
\label{sec:intro}

\begin{figure*}[t]
	\centering
	\includegraphics[width=1\linewidth]{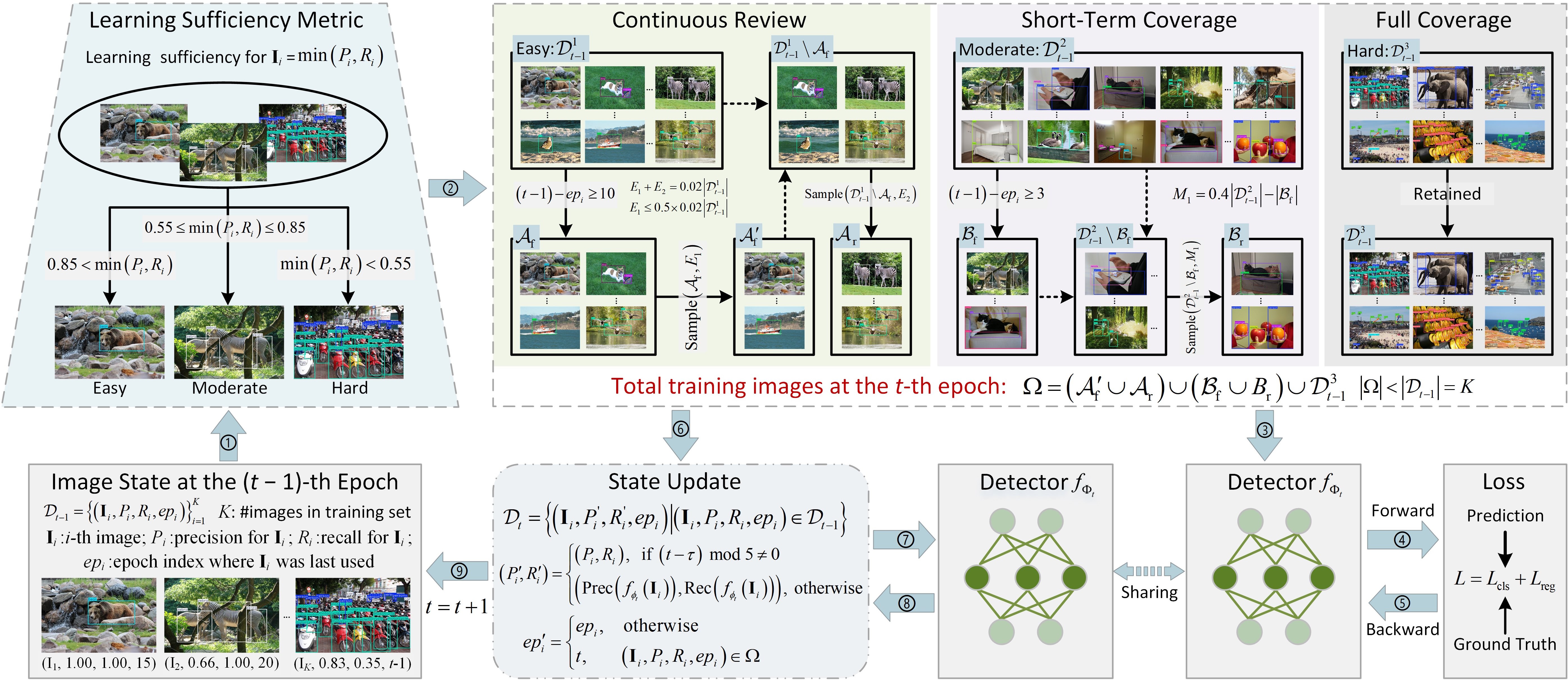}
	\caption{
		Overview of the proposed Anti-Forgetting Sampling Strategy (AFSS) at the $t$-th epoch of training.
	}
	\label{fig2}
\end{figure*}
The workflow of the Anti-Forgetting Sampling Strategy (AFSS) at epoch $t$ is illustrated in Fig.~\ref{fig2}. Given the learning states inherited from the $(t-1)$-th epoch, AFSS first evaluates the learning sufficiency of each training image using precision and recall, and assigns it to one of three difficulty levels: \textit{easy}, \textit{moderate}, or \textit{hard}.
Easy images that have already been well learned are involved with minimal frequency and periodically revisited to avoid catastrophic forgetting. Moderate images are involved with a medium frequency while maintaining short-term full coverage, ensuring each image reappears at least once within three epochs. Hard images are kept fully involved throughout training to guarantee sufficient optimization on challenging cases.
The resulting subset of images is then used to train the YOLO detector at the $t$-th epoch. After training, AFSS updates the learning states of all images, enabling the scheduler to dynamically adjust their involvement in the next epoch.
\subsection{Learning Sufficiency Metric}
To assess whether a training image has been sufficiently learned, AFSS introduces a \textbf{learning sufficiency metric} that jointly considers detection precision and recall. In object detection, an image is reliably learned only when the detector can both correctly classify all objects and localize them completely. We therefore define the learning sufficiency of training image $\mathbf{I}_i$ as:
\begin{equation}
	\text{Learning sufficiency for } \mathbf{I}_i = \min(P_i, R_i).
\end{equation}
where $P_i$ and $R_i$ denote detection precision and recall for $\mathbf{I}_i$. This formulation emphasizes the weaker dimension: an image is regarded as insufficiently learned whenever either classification or localization remains unreliable.

This metric offers several advantages. It reflects task-level learning quality rather than noisy per-iteration losses, aligns directly with the detection objective, and adds almost no computational overhead since modern YOLO detectors already compute these statistics. Overall, it provides a stable, interpretable, and efficient measure of image-level learning progress.
Based on this criterion, AFSS stratifies images into three difficulty levels according to their sufficiency scores:
\begin{equation}
	\label{eq:image-levels}
	\begin{cases}
		\text{Easy}, & \text{if } \min(P_i, R_i) > 0.85, \\
		\text{Moderate}, & \text{if } 0.55 \le \min(P_i, R_i) \le 0.85, \\
		\text{Hard}, & \text{if } \min(P_i, R_i) < 0.55.
	\end{cases}
\end{equation}

Easy images correspond to those the detector already handles confidently.  
Moderate images exhibit partial stability and still require refinement.  
Hard images remain challenging, often due to occlusion, or small object size.

For the $t$-th training epoch, AFSS determines the difficulty level of each training image primarily based on its state from the $(t\!-\!1)$-th epoch. 
AFSS uses the state dictionary $\mathcal{D}_{t-1}$, which records for each image its precision, recall, and last-used epoch from the previous training. 
According to the learning sufficiency metric, AFSS partitions all training images into three subsets:
the \textit{easy subset} $\mathcal{D}_{t-1}^1$,
the \textit{moderate subset} $\mathcal{D}_{t-1}^2$,
and the \textit{hard subset} $\mathcal{D}_{t-1}^3$.

This categorization forms the basis of the subsequent training schedule. AFSS sparsely revisits easy images to prevent forgetting, maintains balanced involvement for moderate images, and allocates more learning effort to hard images that the detector has not yet mastered.

\subsection{Continuous Review}
To mitigate forgetting without repeatedly updating well-learned data, AFSS introduces a \emph{continuous review mechanism} that revisits only a small portion of easy training images at each epoch. Although these training images exhibit high learning sufficiency, omitting them entirely may cause gradual degradation of previously acquired representations.

At the beginning of the $t$-th epoch , AFSS retrieves the easy-training-image set from the previous state dictionary, denoted as $\mathcal{D}_{t-1}^{1}$. 
Each record $(\mathbf{I}_i, P_i, R_i, ep_i)$ stores the training image, its precision, recall, and the last epoch in which it appeared.
Easy training images that have been unused for at least ten epochs are identified as:
\begin{equation}
	\mathcal{A}_{\mathrm{f}} = 
	\left\{\, (\mathbf{I}_i, P_i, R_i, ep_i) \in \mathcal{D}_{t-1}^{1} \mid t - 1 - ep_i \ge 10 \,\right\}.
\end{equation}

AFSS constructs the easy-training-image subset for the current epoch through a two-stage process controlled by $E_1$ and $E_2$.
First, $E_1$ long-unseen training images are sampled from $\mathcal{A}_{\mathrm{f}}$ to form the \emph{forced-review set}:
\begin{equation}
	\mathcal{A}_{\mathrm{f}}' = \texttt{randsample}(\mathcal{A}_{\mathrm{f}},\, E_1),
\end{equation}
These training images are deliberately reintroduced to prevent the detector from forgetting previously mastered but long-unused cases. 
Thus, $E_1$ controls the \emph{strength of the anti-forgetting effect}.

Next, after excluding the forced-review images from the easy pool, AFSS selects $E_2$ training images from the remaining easy set to maintain diversity:
\begin{equation}
	\mathcal{A}_{\mathrm{r}} = 
	\texttt{randsample}\left(\mathcal{D}_{t-1}^{1} \setminus \mathcal{A}_{\mathrm{f}}',\, E_2\right),
\end{equation}
where the set difference $\mathcal{D}_{t-1}^{1} \setminus \mathcal{A}_{\mathrm{f}}'$ denotes the remaining easy training images after removing those already selected for forced review. 
Hence, $E_2$ determines the \emph{diversity level} of easy-image exposure.

The final easy-training-image subset for the $t$-th epoch  is 
$\mathcal{A}_{\mathrm{f}}' \cup \mathcal{A}_{\mathrm{r}}$. AFSS further enforces the scheduling constraints:
\begin{equation}
	E_1 + E_2 = 0.02 \times |\mathcal{D}_{t-1}^{1}|,\quad 
	E_1 \le 0.5 \times 0.02 \times |\mathcal{D}_{t-1}^{1}|.
\end{equation}

These constraints ensure that only $2\%$ of easy training images participate in each epoch, and no more than half of them originate from forced-review instances. The forced-review subset preserves long-term knowledge by periodically revisiting long-unseen training images, while the random subset provides lightweight variability that enhances robustness. Together, they form an efficient and stable rehearsal mechanism that mitigates forgetting, stabilizes optimization, and avoids redundant computation.
\subsection{Short-Term Coverage}
Given the moderate training image set $\mathcal{D}_{t-1}^{2}$ stored in the previous state dictionary, AFSS tracks the most recent training epoch of each image using its recorded value $ep_i$. During epoch $t$, the moderate training images that have not been used for two consecutive epochs form the \textit{forced coverage subset}:
\begin{equation}
	\mathcal{B}_{\mathrm{f}}
	= \left\{
	(\mathbf{I}_i, P_i, R_i, ep_i) \in \mathcal{D}_{t-1}^{2}
	\,\middle|\,
	t - 1 - ep_i \ge 3
	\right\}.
	\label{eq:bf}
\end{equation}
These images are given priority in order to prevent forgetting where prediction confidence remains unstable and to maintain proper alignment between feature representations and category boundaries.

To preserve distributional diversity, AFSS expands this forced coverage subset by selecting additional moderate training images uniformly at random from the remaining pool. The complementary random subset is defined as:
\begin{equation}
	\mathcal{B}_{\mathrm{r}}
	= \texttt{randsample}\!\left(
	\mathcal{D}_{t-1}^{2} \setminus \mathcal{B}_{\mathrm{f}},\,
	M_1
	\right),
	\label{eq:br}
\end{equation}
where the required number of additional images is:
\begin{equation}
	M_1 = 
	0.4\left| \mathcal{D}_{t-1}^{2} \right|
	-
	\left| \mathcal{B}_{\mathrm{f}} \right|.
	\label{eq:m1}
\end{equation}

The final moderate training image set used in epoch $t$ is given by:
\begin{equation}
	\mathcal{B}_{\mathrm{f}} \cup \mathcal{B}_{\mathrm{r}}.
	\label{eq:mt}
\end{equation}
This construction ensures that every moderate training image is revisited within a short temporal window. The forced coverage subset prevents forgetting, while the random subset guarantees sufficient variation in each epoch. Together, they form an efficient and robust self-regularization mechanism that stabilizes the optimization of moderate cases without incurring excessive training cost.
\subsection{State Update}
After the $t$-th epoch, AFSS updates the state dictionary to reflect the detector's current learning status and image usage. The dictionary from the previous epoch is:
\begin{equation}
	\mathcal{D}_{t-1}
	= \{\, (\mathbf{I}_i, P_i, R_i, ep_i) \,\}_{i=1}^{K},
\end{equation}
where $\mathbf{I}_i$ is the $i$-th training image, $P_i$ and $R_i$ denote its precision and recall, and $ep_i$ is the last epoch in which the image was used. After the $t$-th epoch, AFSS generates the updated dictionary:
\begin{equation}
	\mathcal{D}_{t}
	= \{\, (\mathbf{I}_i, P_i', R_i', ep_i') 
	\mid 
	(\mathbf{I}_i, P_i, R_i, ep_i) \in \mathcal{D}_{t-1} \,\}.
\end{equation}

To reduce redundant evaluation, AFSS refreshes precision and recall every five epochs after warm-up. When $(t - \tau) \bmod 5 = 0$,
the performance of each image is updated as:
\begin{equation}
	(P_i', R_i')
	= \bigl(
	\text{Prec}(f_{\phi_t}(\mathbf{I}_i)),\,
	\text{Rec}(f_{\phi_t}(\mathbf{I}_i))
	\bigr),
\end{equation}
otherwise the previous values are retained:
\begin{equation}
	(P_i', R_i') = (P_i, R_i).
\end{equation}

AFSS also updates image usage. Let $\Omega$ denote all training images selected during the $t$-th epoch. For any image in $\Omega$, its usage epoch becomes:
\begin{equation}
	ep_i' =
	\begin{cases}
		t, & (\mathbf{I}_i, P_i, R_i, ep_i) \in \Omega, \\
		ep_i, & \text{otherwise}.
	\end{cases}
\end{equation}

The set $\Omega$ is formed by the three AFSS components:
\begin{equation}
	\Omega 
	= (\mathcal{A}_{t}' \cup \mathcal{A}_{\mathrm{r}})
	\cup 
	(\mathcal{B}_{\mathrm{f}} \cup \mathcal{B}_{\mathrm{r}})
	\cup 
	\mathcal{D}_{t-1}^{3},
	\qquad
	|\Omega| < K.
\end{equation}

Finally, the updated dictionary is propagated to the next epoch.
This continuous update provides AFSS with stable learning-state estimates and accurate usage history, enabling adaptive image scheduling across training.

\newcommand{\subcap}[1]{%
	\parbox{\linewidth}{\noindent\justifying #1}\\[4pt]%
}

\section{Experiments}
In this section, we evaluate AFSS across multiple YOLO models and standard benchmarks. We first describe the datasets and metrics, then present implementation details, main results, and ablation studies.
\begin{table*}[t]  
	\centering
	\fontsize{8.2pt}{8pt}\selectfont
	\setlength{\tabcolsep}{8.2pt}
	\renewcommand{\arraystretch}{1.1}
	\caption{Accuracy and training efficiency of different models with and without AFSS on MS COCO 2017 and PASCAL VOC 2007 datasets, evaluated on two RTX 4090 GPUs (24 GB each). All subsequent results are obtained under the same hardware configuration.}
		\vspace{-2mm}
	\label{tab1}
	\begin{tabular}{c ccc c ccc}
		\toprule
		\multirow{2}{*}{\textbf{Method}} & \multicolumn{3}{c}{\textbf{MS COCO 2017}} & & \multicolumn{3}{c}{\textbf{PASCAL VOC 2007}} \\
		\cmidrule(lr){2-4} \cmidrule(lr){6-8}
		& \textbf{AP} $\uparrow$ & \textbf{Training time (h)} $\downarrow$ & \textbf{Speedup ($\times$)} $\uparrow$ & & \textbf{mAP} $\uparrow$ & \textbf{Training time (h)} $\downarrow$ & \textbf{Speedup ($\times$)} $\uparrow$ \\
		\midrule
		YOLOv8n & 37.3 & 30.4 & -- & & 75.9 & 4.5 & -- \\
		\rowcolor{afssrow}
		YOLOv8n + AFSS & 37.4 & 21.2 & 1.43$\times$ & & 76.0 & 2.8 & 1.60$\times$ \\
		YOLOv10n         & 38.5 & 39.9 & --    & & 76.1 & 5.6 & --    \\
		\rowcolor{afssrow}
		YOLOv10n + AFSS  & 38.6 & 27.4 & 1.45$\times$ & & 76.2 & 3.5 & 1.60$\times$ \\
		YOLO11n         & 39.5 & 32.7 & --    & & 76.3 & 5.2 & --    \\
		\rowcolor{afssrow}
		YOLO11n + AFSS  & 39.6 & 22.1 & 1.47$\times$ & & 76.3 & 3.2 & 1.62$\times$ \\
		YOLO12n         & 40.6 & 43.5 & --    & & 76.6 & 7.3 & --    \\
		\rowcolor{afssrow}
		YOLO12n + AFSS  & 40.6 & 29.2 & 1.48$\times$ & & 76.7 & 4.5 & 1.62$\times$ \\
		\midrule
		
		YOLOv8s          & 45.0 & 42.4 & --    & & 81.1 & 6.7 & --    \\
		\rowcolor{afssrow}
		YOLOv8s + AFSS   & 45.1 & 27.6 & 1.53$\times$ & & 81.2 & 4.1 & 1.63$\times$ \\
		
		YOLOv10s         & 46.3 & 54.2 & --    & & 81.4 & 8.0 & --    \\
		\rowcolor{afssrow}
		YOLOv10s + AFSS  & 46.4 & 35.1 & 1.54$\times$ & & 81.4 & 4.9 & 1.63$\times$ \\
		
		YOLO11s         & 47.0 & 43.9 & --    & & 81.7 & 6.9 & --    \\
		\rowcolor{afssrow}
		YOLO11s + AFSS  & 47.2 & 28.4 & 1.54$\times$ & & 81.8 & 4.2 & 1.64$\times$ \\
		YOLO12s         & 48.0 & 64.6 & --    & & 82.1 & 10.2 & --    \\
		\rowcolor{afssrow}
		YOLO12s + AFSS  & 48.1 & 41.5 & 1.55$\times$ & & 82.2 & 6.2 & 1.64$\times$ \\
		\midrule
		YOLOv8m          & 50.2 & 76.5 & --    & & 83.5 & 11.9 & --    \\
		\rowcolor{afssrow}
		YOLOv8m + AFSS   & 50.3 & 47.7 & 1.60$\times$ & & 83.7 & 7.2 & 1.65$\times$ \\
		YOLOv10m         & 51.1 & 88.2 & --    & & 83.7 & 13.4 & --    \\
		\rowcolor{afssrow}
		YOLOv10m + AFSS  & 51.2 & 54.8 & 1.60$\times$ & & 83.9 & 8.1 & 1.65$\times$ \\
		YOLO11m         & 51.5 & 81.6 & --    & & 84.0 & 12.8 & --    \\
		\rowcolor{afssrow}
		YOLO11m + AFSS  & 51.7 & 50.6 & 1.61$\times$ & & 84.1 & 7.7 & 1.66$\times$ \\
		YOLO12m         & 52.5 & 111.3 & --    & & 84.2 & 17.5 & --    \\
		\rowcolor{afssrow}
		YOLO12m + AFSS  & 52.6 & 68.7 & 1.62$\times$ & & 84.4 & 10.5 & 1.66$\times$ \\
		\midrule
		YOLOv8l          & 52.9 & 111.8 & --    & & 85.1 & 17.5 & --    \\
		\rowcolor{afssrow}
		YOLOv8l + AFSS   & 53.0 & 67.8 & 1.64$\times$ & & 85.1 & 10.5 & 1.66$\times$ \\
		YOLOv10l         & 53.2 & 125.6 & --    & & 85.3 & 19.2 & --    \\
		\rowcolor{afssrow}
		YOLOv10l + AFSS  & 53.3 & 75.7 & 1.65$\times$ & & 85.5 & 11.5 & 1.66$\times$ \\
		YOLO11l         & 53.4 & 102.2 & --    & & 85.6 & 16.0 & --    \\
		\rowcolor{afssrow}
		YOLO11l + AFSS  & 53.4 & 61.6 & 1.65$\times$ & & 85.7 & 9.6 & 1.66$\times$ \\
		YOLO12l         & 53.7 & 168.7 & --    & & 85.8 & 23.9 & --    \\
		\rowcolor{afssrow}
		YOLO12l + AFSS  & 53.8 & 101.5 & 1.66$\times$ & & 85.8 & 14.3 & 1.67$\times$ \\
		\midrule
		YOLOv8x          & 53.9 & 168.2 & --    & & 85.4 & 26.1 & --    \\
		\rowcolor{afssrow}
		YOLOv8x + AFSS   & 54.1 & 100.2 & 1.67$\times$ & & 85.5 & 15.5 & 1.68$\times$ \\
		YOLOv10x         & 54.4 & 170.2 & --    & & 85.7 & 26.3 & --    \\
		\rowcolor{afssrow}
		YOLOv10x + AFSS  & 54.4 & 101.4 & 1.67$\times$ & & 85.8 & 15.6 & 1.68$\times$ \\
		YOLO11x         & 54.7 & 161.6 & --    & & 85.9 & 25.2 & --    \\
		\rowcolor{afssrow}
		YOLO11x + AFSS  & 54.9 & 96.1 & 1.68$\times$ & & 86.0 & 14.9 & 1.69$\times$ \\
		YOLO12x         & 55.2 & 260.6 & --    & & 86.2 & 40.3& --    \\
		\rowcolor{afssrow}
		YOLO12x + AFSS  & 55.4 & 154.8 & 1.68$\times$ & & 86.4 & 23.8 & 1.69$\times$ \\
		\bottomrule
	\end{tabular}
\end{table*}

\begin{table*}[t]  
	\centering
	\fontsize{8.2pt}{8pt}\selectfont
	\setlength{\tabcolsep}{7pt}
	\renewcommand{\arraystretch}{1.1}
	\caption{Accuracy and training efficiency of different models with and without AFSS on DOTA-v1.0 and DIOR-R datasets.}
		\vspace{-2mm}
	\label{tab2}
	\begin{tabular}{c ccc c ccc}
		\toprule
		\multirow{2}{*}{\textbf{Method}} & \multicolumn{3}{c}{\textbf{DOTA-v1.0}} & & \multicolumn{3}{c}{\textbf{DIOR-R}} \\
		\cmidrule(lr){2-4} \cmidrule(lr){6-8}
		& \textbf{mAP} $\uparrow$ & \textbf{Training time (h)} $\downarrow$ & \textbf{Speedup ($\times$)} $\uparrow$ & & \textbf{mAP} $\uparrow$ & \textbf{Training time (h)} $\downarrow$ & \textbf{Speedup ($\times$)} $\uparrow$ \\
		\midrule
		YOLOv8n-OBB        & 78.0 & 94.1 & --      & & 79.9 & 5.1 & -- \\
		\rowcolor{afssrow}
		YOLOv8n-OBB + AFSS & 78.1 & 57.5 & 1.63$\times$ & & 79.9 & 3.1 & 1.64$\times$ \\
		YOLO11n-OBB        & 78.4 & 89.3 & --      & & 80.2 & 5.0 & -- \\
		\rowcolor{afssrow}
		YOLO11n-OBB + AFSS & 78.4 & 53.8 & 1.65$\times$ & & 80.3 & 3.0 & 1.66$\times$ \\
		\midrule
		YOLOv8s-OBB        & 79.5 & 106.2 & --     & & 80.5 & 8.8 & -- \\
		\rowcolor{afssrow}
		YOLOv8s-OBB + AFSS & 79.5 & 63.6  & 1.66$\times$ & & 80.6 & 5.3 & 1.66$\times$ \\
		YOLO11s-OBB        & 79.5 & 102.9 & --     & & 80.6 & 7.5 & -- \\
		\rowcolor{afssrow}
		YOLO11s-OBB + AFSS & 79.6 & 61.5  & 1.67$\times$ & & 80.7 & 4.5 & 1.66$\times$ \\
		\midrule
		YOLOv8m-OBB        & 80.5 & 115.6 & --     & & 81.5 & 11.2 & -- \\
		\rowcolor{afssrow}
		YOLOv8m-OBB + AFSS & 80.8 & 69.2  & 1.67$\times$ & & 81.6 & 6.7  & 1.67$\times$ \\
		YOLO11m-OBB        & 80.9 & 112.8 & --     & & 82.0 & 10.9 & -- \\
		\rowcolor{afssrow}
		YOLO11m-OBB + AFSS & 80.9 & 67.1  & 1.68$\times$ & & 82.2 & 6.5  & 1.67$\times$ \\
		\midrule
		YOLOv8l-OBB        & 80.7 & 165.9 & --     & & 82.5 & 15.4 & -- \\
		\rowcolor{afssrow}
		YOLOv8l-OBB + AFSS & 80.9 & 98.5  & 1.68$\times$ & & 82.5 & 9.2  & 1.67$\times$ \\
		
		YOLO11l-OBB        & 81.0 & 150.0 & --     & & 82.7 & 13.7 & -- \\
		\rowcolor{afssrow}
		YOLO11l-OBB + AFSS & 81.0 & 89.1  & 1.68$\times$ & & 82.8 & 8.1  & 1.69$\times$ \\
		\midrule
		YOLOv8x-OBB        & 81.4 & 246.2 & --     & & 83.6 & 22.3 & -- \\
		\rowcolor{afssrow}
		YOLOv8x-OBB + AFSS & 81.4 & 145.0 & 1.69$\times$ & & 83.6 & 13.1  & 1.70$\times$ \\
		
		YOLO11x-OBB        & 81.3 & 242.2 & --     & & 83.6 & 21.7 & -- \\
		\rowcolor{afssrow}
		YOLO11x-OBB + AFSS & 81.4 & 143.1 & 1.69$\times$ & & 83.7 & 12.7  & 1.70$\times$ \\
		\bottomrule
	\end{tabular}
\end{table*}

\subsection{Datasets and Evaluation Protocols}
We evaluate AFSS on four widely used object detection benchmarks covering both natural images and remote sensing images. For natural images, we use MS COCO 2017~\cite{lin2014microsoft} and PASCAL VOC 2007~\cite{everingham2015pascal}, two standard datasets commonly used to assess modern detectors. For remote sensing images, we adopt DOTA-v1.0~\cite{xia2018dota} and DIOR-R~\cite{cheng2022anchor}, which provide large-scale imagery with diverse object categories, arbitrary orientations, and cluttered backgrounds.
Following standard evaluation protocols, we report AP (AP@[.5:.95]) on MS COCO 2017, and mAP at the IoU threshold of 0.5 on PASCAL VOC 2007, DOTA-v1.0, and DIOR-R.
\subsection{Implementation Details}
All experiments are conducted using Ultralytics YOLO framework on two RTX 4090 GPUs (24 GB each). The default batch size is 64 (32 images per GPU), and is reduced for larger models to avoid memory overflow.
Input resolutions are configured according to the characteristics of each dataset: 
640$\times$640 for MS COCO 2017 and PASCAL VOC 2007, 1024$\times$1024 for DOTA-v1.0, and 
800$\times$800 for DIOR-R. 
All models follow the standard training schedules used in official YOLO implementations: 
600 epochs for MS COCO 2017 and 
300 epochs for PASCAL VOC 2007, DOTA-v1.0, and DIOR-R. 
Unless otherwise specified, all hyperparameters and optimization settings strictly follow the default configurations provided by the Ultralytics YOLO repository.

\subsection{Main Results}
We evaluate AFSS on MS COCO 2017 and PASCAL VOC 2007 using YOLOv8, YOLOv10, YOLOv11, and YOLOv12 across all standard model scales (n/s/m/l/x). As shown in Table~\ref{tab1}, AFSS achieves more than \textbf{$1.43\times$} training speedup while maintaining or improving accuracy. This demonstrates that AFSS removes substantial redundant computation without compromising detector quality.
We further validate AFSS on remote sensing detection by integrating it into YOLOv8-OBB and YOLOv11-OBB. As reported in Table~\ref{tab2}, AFSS achieves more than \textbf{$1.63\times$} training speedup together with consistent mAP gains on DOTA-v1.0 and DIOR-R. These results show that AFSS is especially effective in dense and complex scenes. 

Overall, AFSS offers a architecture-agnostic way to accelerate detector training while maintaining high accuracy. Its consistent gains across natural and remote sensing datasets highlight AFSS as a practical solution to the rising computational cost of modern detection models.

\begin{table}[t]
	\centering
	\caption{Comparison of different training methods.}
	\vspace{-2mm}
	\resizebox{0.98\linewidth}{!}{
		\begin{tabular}{lccc}
			\toprule
			\textbf{Method} & \textbf{AP} $\uparrow$ & \textbf{Training time (h)} $\downarrow$ & \textbf{Speedup} $\uparrow$ \\
			\midrule
			YOLO11s (Baseline) & 47.0 & 43.9 & -- \\
			+ Curriculum Learning & 43.7 & 32.5 & 1.35$\times$ \\
			+ Self-Paced Learning & 44.5 & 33.6 & 1.30$\times$ \\
			+ Data Pruning & 40.5 & 31.8 & 1.38$\times$ \\
			+ Dataset Distillation & 35.6 & 29.2 & 1.50$\times$ \\
			\rowcolor{gray!10} + AFSS (Ours) & 47.2 & 28.4 & 1.54$\times$ \\
			\bottomrule
	\end{tabular}}
	\label{tab3}
\end{table}

\begin{table}[t] 
	\centering
	\fontsize{7pt}{6pt}\selectfont
	\renewcommand{\arraystretch}{1.1}
	\setlength{\tabcolsep}{8pt}
	\caption{Ablation of AFSS core components using YOLOv11s as the baseline. 
		LSM: Learning Sufficiency Metric; CR: Continuous Review; STC: Short-Term Coverage; SU: State Update.}
	\vspace{-2mm}
	\label{tab4}
	\resizebox{\columnwidth}{!}{
		\begin{tabular}{cccccc}
			\toprule
			\textbf{LSM} & \textbf{CR} & \textbf{STC} & \textbf{SU} & \textbf{AP} $\uparrow$ & \textbf{Speedup} $\uparrow$  \\
			\midrule
			&  &  &  & 47.0 & -- \\
			\checkmark &  &  &  & 44.8 & 1.45$\times$ \\
			\checkmark & \checkmark &  &  & 45.5 & 1.34$\times$ \\
			\checkmark &  & \checkmark &  & 46.6 & 1.31$\times$ \\
			\checkmark & \checkmark & \checkmark &  & 47.2 & 1.26$\times$ \\
			\rowcolor{afssrow}
			\checkmark & \checkmark & \checkmark & \checkmark & 47.2 & 1.54$\times$ \\
			\bottomrule
		\end{tabular}
	}
\end{table}

\begin{table*}[t]
	
	\caption{Ablation studies on learning sufficiency metric, continuous review interval,
		short-term coverage interval, and state update interval.}
	\label{tab5}
	\vspace{-1mm}
	\fontsize{7.3pt}{7pt}\selectfont
	\renewcommand{\arraystretch}{1.08}
	\setlength{\tabcolsep}{5.8pt}
	
	\begin{minipage}[t]{0.30\textwidth}\centering
		\parbox{0.82\linewidth}{\noindent\justifying
			(a) Our learning sufficiency metric \textbf{min(Prec, Rec)} vs. other metrics, where Prec and Rec are Precision and Recall.}
		\\[2.3pt]
		\begin{tabular}{lcc}
			\toprule
			\textbf{Metrics} & \textbf{AP} $\uparrow$ & \textbf{Speedup} $\uparrow$\\
			\midrule
			Loss-based      & 46.0 & 1.52$\times$\\
			Gradient-based  & 46.9 & 1.45$\times$\\
			F1 score-based  & 46.6 & 1.51$\times$\\
			\rowcolor{afssrow}
			Our metric      & 47.2 & 1.54$\times$\\
			\bottomrule
		\end{tabular}
	\end{minipage}%
	\hspace{-2.8mm}
	\begin{minipage}[t]{0.233\textwidth}\centering
		\parbox{0.88\linewidth}{\noindent\justifying
			(b) Effect of \textbf{continuous review interval} on the performance of accuracy and training time.}
		\\[2pt]
		\centering
		\begin{tabular}{ccc}
			\toprule
			\textbf{Interval} & \textbf{AP} $\uparrow$ & \textbf{Speedup} $\uparrow$\\
			\midrule
			5  & 47.2 & 1.49$\times$\\
			\rowcolor{afssrow}
			10 & 47.2 & 1.54$\times$\\
			15 & 45.7 & 1.58$\times$\\
			20 & 44.8 & 1.63$\times$\\
			\bottomrule
		\end{tabular}
	\end{minipage}%
	\begin{minipage}[t]{0.233\textwidth}\centering
		\parbox{0.88\linewidth}{\noindent\justifying
			(c) Effect of \textbf{short-term coverage interval} on the performance of accuracy and training time.}
		\\[2pt]
		\centering
		\begin{tabular}{ccc}
			\toprule
			\textbf{Interval} & \textbf{AP} $\uparrow$ & \textbf{Speedup} $\uparrow$\\
			\midrule
			2 & 47.2 & 1.53$\times$\\
			\rowcolor{afssrow}
			3 & 47.2 & 1.54$\times$\\
			4 & 46.0 & 1.55$\times$\\
			5 & 44.2 & 1.56$\times$\\
			\bottomrule
		\end{tabular}
	\end{minipage}%
	\begin{minipage}[t]{0.233\textwidth}\centering
		\parbox{0.88\linewidth}{\noindent\justifying
			(d) Effect of sample \textbf{state update interval} on the performance of accuracy and training time.}
		\\[2pt]
		\centering
		\begin{tabular}{ccc}
			\toprule
			\textbf{Interval} & \textbf{AP} $\uparrow$ & \textbf{Speedup} $\uparrow$\\
			\midrule
			1  & 47.2 & 1.26$\times$\\
			\rowcolor{afssrow}
			5  & 47.2 & 1.54$\times$\\
			10 & 45.8 & 1.72$\times$\\
			15 & 43.7 & 1.93$\times$\\
			\bottomrule
		\end{tabular}
	\end{minipage}
	
\end{table*}

\subsection{Ablation Studies}
We perform ablation studies on MS COCO 2017 using YOLO11s as the baseline, which provides a lightweight yet representative configuration of the YOLO family.
\subsubsection{Comparison with Other Training Strategies}
Table~\ref{tab3} compares AFSS with common training methods on MS COCO 2017 using YOLOv11s. Curriculum learning and self-paced learning achieve a training speedup of 1.35$\times$ and 1.30$\times$, respectively, but both methods introduce fixed easy-to-hard schedules. As a result, training images near the decision boundary receive insufficient supervision, leading to clear accuracy degradation (43.7 AP and 44.5 AP). Data pruning further accelerates training with a speedup of 1.38$\times$, yet removing samples yields biased gradients and forgetting, reducing accuracy to 40.5 AP. Dataset distillation achieves a slightly higher training speedup (1.50$\times$), but the lack of sample diversity in the synthetic set causes the largest performance collapse (35.6 AP). AFSS provides the best overall performance among all methods. It achieves the highest training speedup of 1.54$\times$, while also improving accuracy from 47.0 to 47.2 AP. This demonstrates that dynamically adjusting sample participation and revisiting informative samples when necessary is more effective than fixed curricula, static pruning, or synthetic compression. Overall, AFSS achieves consistent training speedup without sacrificing coverage or stability, offering a practical solution that improves training efficiency while preserving full model performance.

\subsubsection{Effect of Individual Modules}
Table~\ref{tab4} reports how each part of AFSS contributes to the final performance.
Using only the Learning Sufficiency Metric (LSM) achieves a training speedup of 1.45$\times$ but also causes a noticeable accuracy drop, showing that filtering samples by difficulty alone is not sufficient.
Adding Continuous Review (CR) improves the performance further, achieving a training speedup of 1.34$\times$ and raising the accuracy to 45.5 AP.
Incorporating Short-Term Coverage (STC) leads to further accuracy improvements, with a training speedup of 1.31$\times$ and an accuracy of 46.6 AP.
When all components, LSM, CR, and STC, are used together, the performance improves to 47.2 AP with a training speedup of 1.26$\times$.
Finally, with the addition of State Update (SU), AFSS achieves its best performance, with 47.2 AP and the highest training speedup of 1.54$\times$.
Taken together, these results show that AFSS functions as a cohesive system where each component contributes to either stability or efficiency.
\begin{figure}[t]
	\centering
	\includegraphics[width=0.98\linewidth]{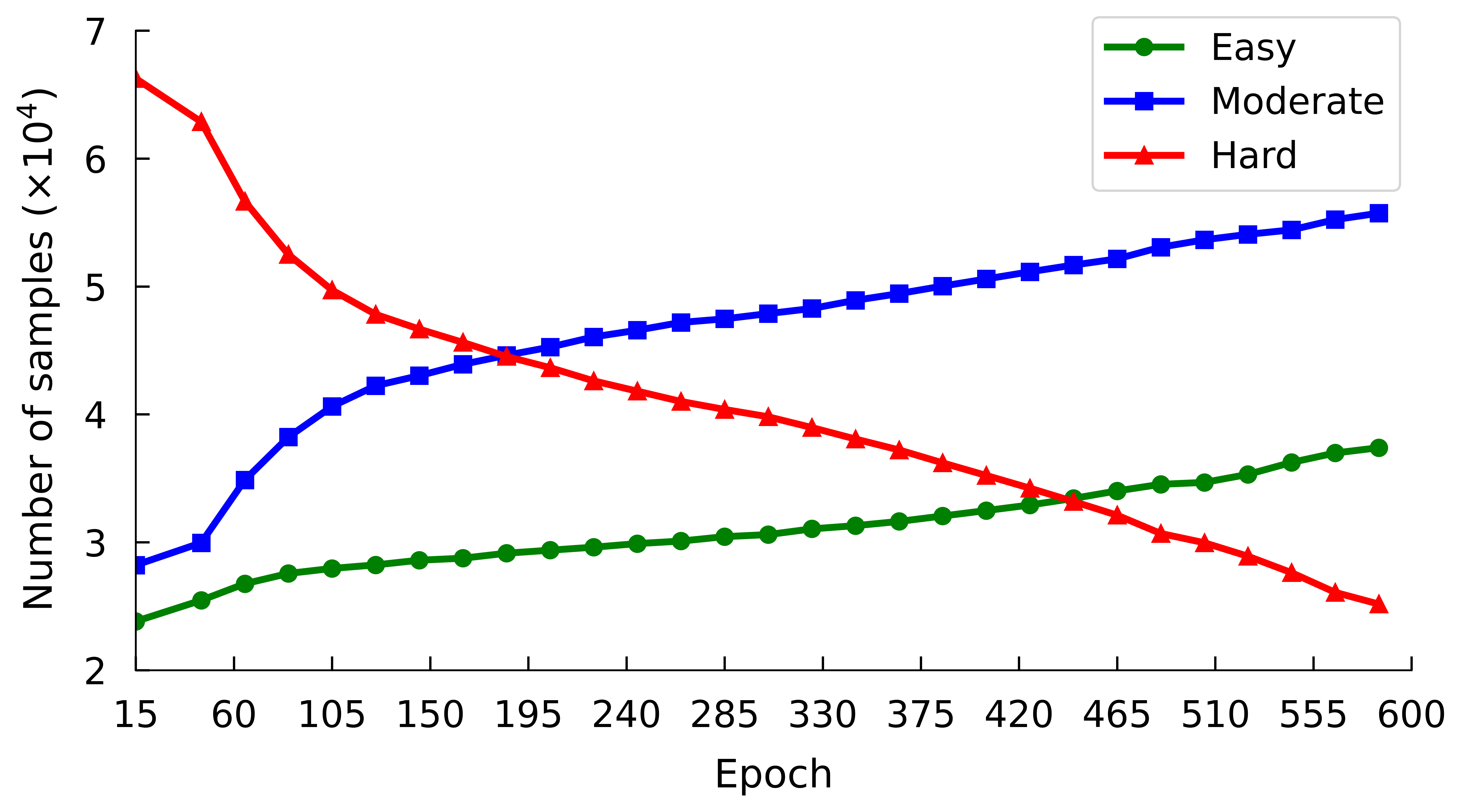}
	\vspace{-2mm}
	\caption{
		Changes in the number of samples at easy, moderate, and hard levels during the training of YOLO11s with AFSS.
	}
	\label{fig3}
\end{figure}
\begin{figure}[t]
	\centering
	\includegraphics[width=0.98\linewidth]{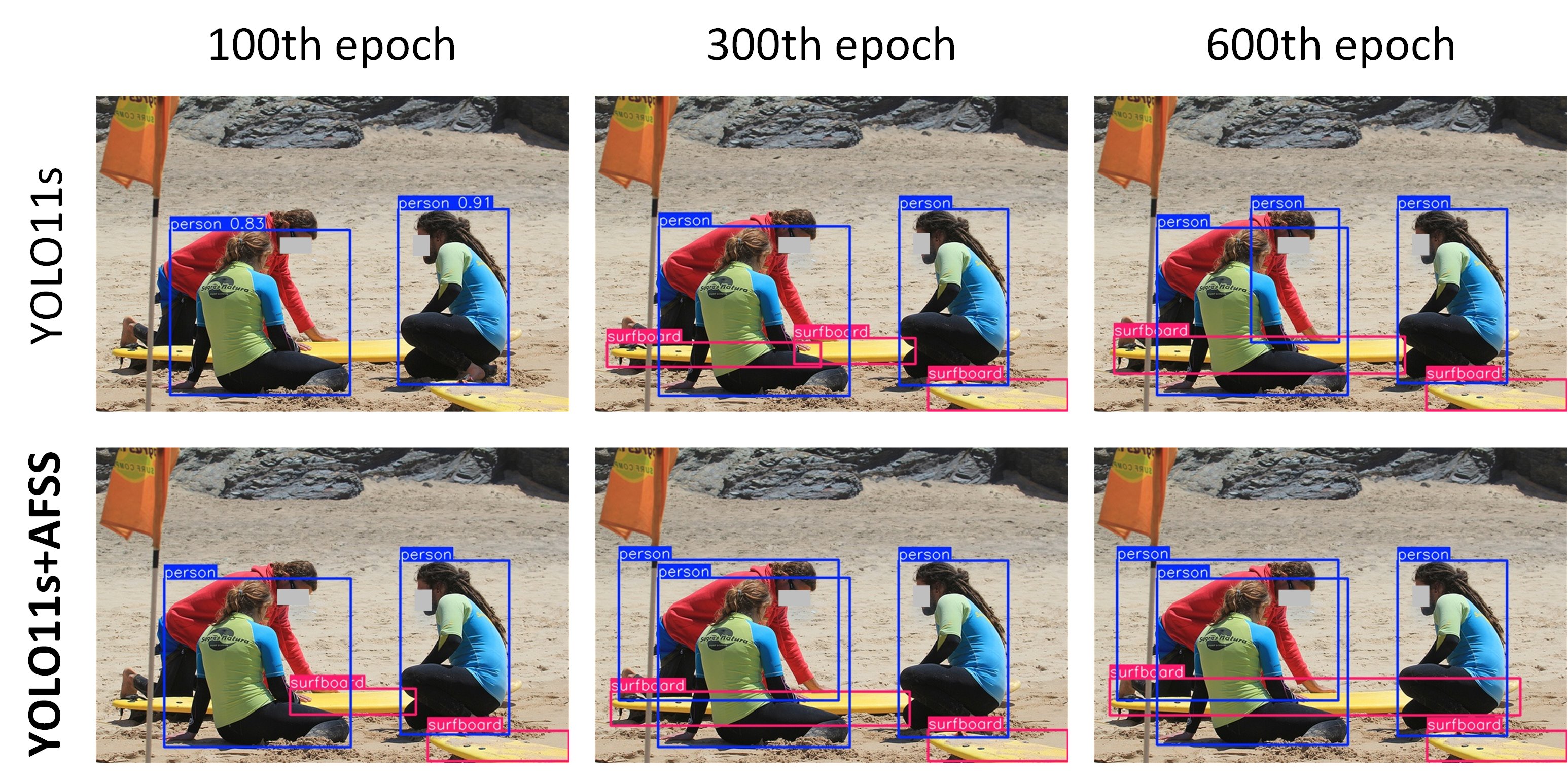}
	\caption{
		An example illustrating the learning performance of YOLO11s and YOLO11s+AFSS at the 100th, 300th, and 600th training epochs on the hard image.
	}
	\label{fig4}
\end{figure}
\subsubsection{Comparison of Learning Sufficiency Metrics}
Table~\ref{tab5} (a) compares several formulations of the learning sufficiency metric.
The loss-based and gradient-based metrics reduce training time but do not effectively distinguish between under-learned and fully learned samples. The loss-based metric saturates early, resulting in an accuracy of 46.0 AP, while the gradient-based metric improves accuracy slightly to 46.9 AP but also increases training time, leading to a training speedup of 1.45$\times$.
The F1 score-based metric also saturates quickly, reaching only 46.6 AP, with a training speedup of 1.51$\times$.
In contrast, our proposed $\min(\text{Prec}, \text{Rec})$ metric focuses on the weaker prediction score, effectively highlighting samples that are still insufficiently learned. By avoiding unnecessary training, our approach achieves the best performance, reaching 47.2 AP with a training speedup of 1.54$\times$.
\subsubsection{Analysis of Interval Design}
For continuous review interval, an interval of 10 provides the best balance between accuracy and efficiency, achieving 47.2 AP with a training speedup of 1.54$\times$. Smaller intervals (such as 5) revisit easy samples too frequently, while larger intervals (such as 15 or 20) delay reinforcement, leading to a reduction in accuracy.
For short-term coverage interval, an interval of 3 offers stable performance. Shorter intervals (such as 2) cause unnecessary revisits of moderate samples, while larger intervals (such as 5) weaken coverage and reduce generalization.
For state update interval, updating every 5 epochs provides the optimal trade-off. More frequent updates (such as every 1 epoch) increase computational cost, while less frequent updates (such as 10 or 15 epochs) produce outdated sufficiency estimates, which degrade accuracy.
Overall, moderately set intervals allow AFSS to maintain accurate sample assessments and consistent supervision, ensuring efficient training without sacrificing performance.

\subsubsection{Accuracy Improvement Analysis}
Fig.~\ref{fig3} presents the evolution of image difficulty over training. From the results, we observe a clear trend: the number of hard images steadily decreases, while easy and moderate images increase, indicating that the model learns challenging cases more quickly when AFSS is applied.
Fig.~\ref{fig4} provides a qualitative comparison under the same number of epochs. It shows that the model trained with AFSS detects hard instances more reliably than the baseline, which still produces missed or incorrect predictions. These results collectively indicate that AFSS reduces redundant updates and directs computation toward informative images, enabling faster learning and stronger final performance.

\section{Conclusions}
We introduce AFSS, an adaptive sampling strategy that enhances YOLO training by minimizing redundant image processing and reallocating computation to informative images. By adjusting exposure based on learning sufficiency, AFSS significantly reduces training costs while preserving critical knowledge. On widely used natural image benchmarks and remote sensing datasets, AFSS achieves over 1.43$\times$ training speedup for YOLO-based detectors, while  improving accuracy. These results show that effective detector training relies on informative samples, not repeated full-dataset processing. AFSS integrates seamlessly with YOLO-based models.

{\sloppy
	\noindent\textbf{Acknowledgements}: This work was supported in part by the National Natural Science Foundation of China (62501478 and 62376223), and by the Fundamental Research Funds for the Central Universities (D5000250066).\par
}
{
    \small
    \bibliographystyle{ieeenat_fullname}
    \bibliography{refs}
}


\end{document}